\documentclass{article}

    \usepackage[final]{neurips_2022}

\setcitestyle{numbers,square}
\usepackage[utf8]{inputenc} 
\usepackage[T1]{fontenc}    
\usepackage{hyperref}       
\usepackage{url}            
\usepackage{booktabs}       
\usepackage{amsfonts}       
\usepackage{nicefrac}       
\usepackage{microtype}      
\usepackage{xcolor}         
\usepackage{amsmath}
\usepackage{bm}
\usepackage{amssymb}
\usepackage{graphicx}
\usepackage{subfigure}
\usepackage{wrapfig}

\title{Deep Surrogate Docking: Accelerating Automated Drug Discovery with Graph Neural Networks}

\author{%
  Ryien Hosseini\thanks{Additional affiliation: Department of Electrical Engineering and Computer Science, University of Michigan, Ann Arbor, MI 48109} \\
  Argonne Leadership Computing Facility\\
  Argonne National Laboratory\\
  Lemont, IL 60439 \\
  \texttt{rhosseini@anl.gov} \\
  \And
  Filippo Simini \\
  Argonne Leadership Computing Facility\\
  Argonne National Laboratory\\
  Lemont, IL 60439 \\
  \texttt{fsimini@anl.gov} \\
  \And 
  Austin Clyde\thanks{Additional affiliation: Department of Computer Science, University of Chicago, Chicago, IL 60657} \\
  Data Science and Learning Division\\
  Argonne National Laboratory\\
  Lemont, IL 60439 \\
  \texttt{aclyde@anl.gov} \\
  \And
  Arvind Ramanathan \\
  Data Science and Learning Division\\
  Argonne National Laboratory\\
  Lemont, IL 60439 \\
  \texttt{ramanathana@anl.gov} \\
}

\begin{document}

\maketitle

\begin{abstract}
The process of screening molecules for desirable properties is a key step in several applications, ranging from drug discovery to material design. During the process of drug discovery specifically, protein-ligand docking, or \textit{chemical docking}, is a standard \textit{in-silico} scoring technique that estimates the binding affinity of molecules with a specific protein target. Recently, however, as the number of virtual molecules available to test has rapidly grown, these classical docking algorithms have created a significant computational bottleneck. We address this problem by introducing \textit{Deep Surrogate Docking (DSD)}, a framework that applies deep learning-based \textit{surrogate modeling} to accelerate the docking process substantially. DSD can be interpreted as a formalism of several earlier surrogate prefiltering techniques, adding novel metrics and practical training practices. Specifically, we show that graph neural networks (GNNs) can serve as fast and accurate estimators of classical docking algorithms. Additionally, we introduce \textit{FiLMv2}, a novel GNN architecture which we show outperforms existing state-of-the-art GNN architectures, attaining more accurate and stable performance by allowing the model to filter out irrelevant information from data more efficiently. Through extensive experimentation and analysis, we show that the DSD workflow combined with the FiLMv2 architecture provides a 9.496x speedup in molecule screening with a $<3\%$ recall error rate on an example docking task. Our open-source code is available at \href{https://github.com/ryienh/graph-dock}{https://github.com/ryienh/graph-dock}.\end{abstract}

\section{Introduction}
\label{intro}



The discovery of new chemical compounds is critical to many scientific fields, such as the development of new medicines, environmental energy sources, and sustainable agriculture. Previously, discovery of such molecules involved costly \textit{in-vitro} synthesis and testing of several candidate compounds \cite{docking_survey}. However, in recent years, advancement in computational hardware has led to \textit{in-silico} screening of new chemical compounds. Concretely, representations of chemicals stored in virtual libraries are screened for desirability using computational scoring functions, and top candidates are selected for \textit{in-vitro} synthesis, testing, and analysis \cite{emulating_docking_ml, lyu_nature, docking_survey}. One of the primary applications of this process is new drug discovery. A particularly popular method for \textit{in-silico} screening of potential compounds for new drug desirability is \textit{protein-ligand docking}. At a high level, docking software analyzes the conformation and orientation (known together as \textit{pose}) of a candidate molecule into the binding site of a particular \textit{target} \cite{emulating_docking_ml}. By simulating the \textit{pose} of the candidate molecule several times, the algorithm assigns the candidate a \textit{docking score}, which represents the estimated binding affinity of the molecule to the target~\cite{docking_survey}. Top docking candidates \footnote{Oftentimes, the top candidates are first selected and then narrowed down further using domain specific methods such as scaffold analysis. These details are outside of the scope of this work, which is focused on accelerating the \textit{docking} phase of filtering.} (those with the \textit{most negative} docking scores) are then selected for wet-lab synthesis and testing \cite{docking_survey}. Recently, \cite{lyu_nature} demonstrated the viability of this docking-enabled drug discovery at scale. Specifically, they demonstrate docking a virtual library of over 170 million molecules against two target receptors, and, by selecting only the top 44 and 549 chemicals, respectively, yielded effective new potential drugs that target each of these receptors. However, this docking process is computationally expensive, and generally CPU bound \cite{docking_algos, docking_survey}. Indeed, the approach used by \cite{lyu_nature} to dock 138 million molecule against a single target required 43,563 CPU core hours to compute. 

In order to solve this problem, recent research has focused on using deep learning to accelerate the docking process~\cite{emulating_docking_ml}. Several methods attempt to circumvent the classical docking algorithm altogether, either by using generative models to produce \textit{de-novo} candidates with desirable properties~\cite{barking, rl_generative_drugs, denovo_overview, gan_denovo, xiong2021graph}, or by using surrogate models to predict docking scores~\cite{emulating_docking_ml,spfd}. Recently, graph neural networks (GNNs) have attracted increasing attention from this field, as molecules can be concisely represented as graphs~\cite{lim2019predicting, xiong2021graph, emulating_docking_ml}. GNNs are generalizations of common deep learning architectures, such as CNNs and attention networks, and are able to directly operate on graph structured data such as molecules~\cite{gnn_survey}. While current methods show significant promise, they suffer from issues such as 
candidate drug synthesizability issues~\cite{barking} and model inaccuracy~\cite{emulating_docking_ml}. 

In this work, we attempt to address the issues with current deep-learning enabled drug discovery workflows. We do so by (i) proposing a novel \textit{Deep Surrogate Docking} (DSD) workflow, which simplifies and improves the hit-rate of current methods by combining ML-based surrogate docking with classical docking, 
(ii) proposing a novel GNN architecture, \textit{FiLMv2}, which we show outperforms existing state of the art GNN architectures in an example DSD workflow, and (iii) providing additional auxiliary improvements to the surrogate training process that increase the overall hit rate of the model. Theoretical discussion and extensive experimentation demonstrates the viability of both DSD and the FiLMv2 architecture in accurately accelerating the throughput of the chemical docking task. Specifically, we demonstrate a 9.496x computational speedup with $<3\%$ error rate compared to classical docking on the ZINC virtual library when targeting the D$_4$ dopamine receptor. 

Section \ref{methods:spfd} describes the DSD workflow. Concretely, a small randomly sampled subset of the virtual molecule library is run through a classical docking algorithm to generate labels (i.e. ground truth docking scores), which are in turn used to train the surrogate model to predict docking scores given a molecule. The trained surrogate model is then used to perform inference on the full molecule library, and the top scoring molecule candidates are run through the classical docking algorithm. 
The novelties of the DSD formalism with respect to earlier pre-filtering methods include the introduction of refined evaluation metrics and training strategies. 
Our analysis framework indicates our workflow is nearly 10x as fast as the traditional-only docking method with less than 3\% error of detecting top hits. 

Section \ref{methods:filmv2} describes the FiLMv2 architecture, a simple iterative improvement to the existing FiLM architecture which improves model performance in our application. Specifically, the affine transformation used in the FiLM architecture is added to the node representation \textit{after}, rather than before, the non-linearity. This increases the expressiveness of the model by preventing collapse to a single affine transformation. We experimentally demonstrate the improvement of the FiLMv2 architecture compared to other 
GNN architectures in the DSD surrogate model task.

\section{Methods}
\label{methods}
Here, we present the key methods of our paper, including the details of our surrogate model workflow, the FiLMv2 architecture, the  Exponentially Weighted Mean Squared Error Loss and Virtual Node. 

\subsection{Deep Surrogate Docking}
\label{methods:spfd}


At a high level, surrogate models aim to emulate, and in many cases, replace \cite{deep_dock} \cite{memes}, classical docking simulation software described in Sections \ref{intro} and \ref{related_work}. These classical docking methods are generally computationally intensive and limited to CPU \cite{lyu_nature, docking_survey}. Thus, surrogate models can enable an exciting opportunity to extend docking over to the ever-increasing size \cite{zinc, docking_survey} of virtual libraries. However, previous analysis \cite{docking_survey, deep_dock} has shown that predicting exact docking scores as a regression task can be infeasible.
%
In order to address this issue, we propose the \textit{Deep Surrogate Docking} (DSD) framework. In this workflow, \textit{surrogate models} are trained to select top candidate molecules which are then docked using a classical algorithm in order to determine the exact docking score and pose information as standard docking workflow. Existing prefiltering techniques such as those proposed in \cite{dsd_1} and \cite{dsd_2} can be interpreted as special cases of our DSD framework. 
Specifically, we define a molecule as a graph $ m = (V, E)$ where $V$ is a set of vertices and $E$ a set of edges. We define a given virtual library of molecules as $M = \{m_1, m_2, \dots, m_N \}$ with $|M| = N$. We then define a classical docking algorithm as a function $f_D: (V, E) \rightarrow \mathbb{R} $ which maps a given molecule to a scalar docking score \footnote{For simplicity, pose estimations are ignored in this discussion, though in practice a pose estimate can also be output from the model. See Appendix for more details.}. Similarly, we can define a surrogate docking model as a 
function $f_{S}: (V, E) \rightarrow \mathbb{R}$, which attempts to estimate the output of $f_D$. We implement this surrogate model as a novel GNN architecture described in Section \ref{methods:filmv2}. 

In our workflow, we first randomly sample a (typically small) subset of $M$: $D \subseteq M$, with $|D| = \rho N$ 
determined by hyperparameter $\rho$ (See Section \ref{experiments:main_results} for more details). We then run this subset $D$ through the docking algorithm in order to generate labels (docking scores) for each $m \in D$: $Y = \{y_i = f_D(m_i) \}_{i=1}^{|D|}$, where $\forall i \; m_i \in D$. We can then define the training set used for training $f_{S}$ as $T = \{(m^{(i)}, y^{(i)})\}_{i=1}^{|D|}$, 
 where $\forall i \; y_i \in Y$. This train set can optionally further be split into typical train-val-test sets for the purposes of model evaluation (See \ref{experiments:exp_details} for details on our data splits). We emphasize here that in chemical docking, molecules with the \textit{lowest} docking scores correspond to the \textit{most desirable}, and we thus use \textit{lowest} and \textit{best} scores interchangeably.  
This thus coverts the problem of training $f_{S}$ into a fully supervised machine learning task. $f_{S}$ is then trained on $T$ using typical gradient descent methods (See Section \ref{experiments} and Appendix for more details on our training methods). 

Next, given a fully trained surrogate model $f_{S}$, we can perform inference on the full virtual library $M$ in order to generate estimated docking scores $y_{S} \in Y_{S} = \{ f_{S}(m_i) \}_{i=1}^{N}$, where $\forall i \; m_i \in M$. We then define the set $P \subset M$ selecting the top $|P| = \sigma N$ molecules with the lowest estimated docking scores. 
The {\it Data Reduction Factor} $\sigma \ll 1$ is the hyperparamter that determines the fraction of most promising (top scoring) molecules that are docked by $f_S$. These $|P|$ molecules are then input to $f_D$ in order to determine the exact docking score and pose information as standard docking workflow: $Y_{P} = \{ f_{D}(m_i) \}_{i=1}^{|P|}$, where $\forall i \; m_i \in P$. 

We now define, for the purposes of evaluation, the {\it Hit Threshold}, $\zeta$, a third hyperparameter which determines the fraction of molecules which should be recalled from the model. For example, when $\zeta = 0.001$, we are interested in correctly selecting the top 0.1\% of molecules in a given virtual library. Our analysis in Section \ref{experiments} uses $\zeta = 0.001$ and $\zeta = 0.01$, which is consistent with typical practices in standard docking \cite{docking_survey, lyu_nature}.

One benefit of such a workflow is that it easily allows introduction of inductive bias into $f_{S}$ so as to properly rank \textit{only top performing} molecule candidates. Typically, only molecules on the order of the top 0.1\% or less of the total virtual library are selected for additional domain-specific analysis and possible synthesis \cite{lyu_nature, docking_survey, docking_algos}. Thus, biases that aid the model in classifying these top molecules, perhaps at the expense of total performance across all docking scores, is useful to the task. One example of such an inductive bias is the exponentially weighted MSE loss described in Section \ref{methods:wmse}. 
Another benefit of this model is the ability to customize the runtime performance vs. accuracy tradeoff. Specifically, as hyperparameter $\sigma$
is varied, the total error rate with respect to traditional docking and time required for the full DSD is intuitively traded off. Section \ref{experiments:main_results} shows that the intuitive notion that $\sigma$ and model accuracy are positively correlated, holds empirically. 
%

\subsection{FiLMv2 Architecture}
\label{methods:filmv2}

Our proposed model to identify molecules with the lowest docking scores is based on a new GNN architecture called FiLMv2. 
FiLMv2 is a variant of the GNN-FiLM architecture~\cite{gnn_film}, which has been shown to outperform other state-of-the-art architectures, like Graph Convolutional Networks (GCN)~\cite{gcn} and Graph Attention Networks (GAT)~\cite{gat}, on regression tasks on molecular graphs.  
The idea behind the original GNN-FiLM model is to perform feature-wise linear modulation of a message's source representation, $\boldsymbol{h}_{u}^{t}$, using a function of the target representation, $\boldsymbol{h}_{v}^{t}$: 
\begin{align}
  \boldsymbol{h}_{v}^{t+1} & = 
    l\left(
      \sum_{{u} \in \mathcal{N}_v}
        \sigma \left( (\boldsymbol{W}_\gamma \boldsymbol{h}_{v}^{t}) \odot (\boldsymbol{W}_\alpha \boldsymbol{h}_{u}^{t}) + 
        (\boldsymbol{W}_\beta \boldsymbol{h}_{v}^{t})
              \right)
    \right) \label{eq:gnn_filmv1}
\end{align}
where $\boldsymbol{W}_\alpha, \boldsymbol{W}_\beta, \boldsymbol{W}_\gamma$ are learned matrices, $\odot$ is the Hadamard element-wise product, $\mathcal{N}_v$ denotes the neighbors of the target node $v$, and $\sigma$ and $l$ are ReLU nonlinearities. The GNN-FiLM architecture of Eq.~\ref{eq:gnn_filmv1} allows to build dynamic filters based on the target node representation to decide which features of the source representation should be discarded or selected in the updated state of the target node. 
The possibility to apply element-wise modulation of a source's representation based on the target's representation is absent in GCN and GAT architectures, where the target representation is either used as an additional incoming message (GCN) or used to modulate the entire source representation (GAT). 
Our FiLMv2 architecture introduces a simple modification to the original GNN-FiLM Eq.~\ref{eq:gnn_filmv1}, consisting in applying a ReLU nonlinearity to the target representations on both the multiplicative and additive terms: 
\begin{align}
 \boldsymbol{h}_{v}^{t+1} & = 
    l\left(
      \sum_{{u} \in \mathcal{N}_v}
        \left( 
        \text{ReLU}(\boldsymbol{W}_\gamma \boldsymbol{h}_{v}^{t}) \odot 
        (\boldsymbol{W}_\alpha \boldsymbol{h}_{u}^{t}) + 
        \text{ReLU}(\boldsymbol{W}_\beta \boldsymbol{h}_{v}^{t})
        \right)
    \right). 
\label{eq:gnn_filmv2}
\end{align}
The addition of the ReLU nonlinearities enables to perform a more effective selection of the most relevant information from the source representation. Indeed, target features that are mapped by the linear transformation to any negative number are zeroed out by the ReLU activation function and hence can effectively filter out the corresponding entries of the source representation via the element-wise products.   
This filtering mechanism is considerably more accurate and stable than the original GNN-FiLM, where a fine-tuned calibration of the linear transformation's parameters is necessary in order to produce an output close to zero. Moreover, without the activation function, for each different target representation a different set of parameter values must be learnt in order to zero-out a particular entry of the source representation, which is highly inefficient and unstable to changes of the target representation. 
These insights are confirmed by a series of tests performed to assess the effectiveness of various combinations of nonlinearities. In particular, we observe that (see Appendix for details of the experiments): 
(i) the nonlinearity $\sigma$ in Eq.~\ref{eq:gnn_filmv1} can be removed as it does not improve the model's performance; 
(ii) adding a nonlinearity to the source term $(\boldsymbol{W}_\alpha \boldsymbol{h}_{u}^{t})$ does not further improve the model performance; 
(iii) using a tanh instead of ReLU nonlinearity does not bring any improvement with respect to the original GNN-FiLMv1. 
The last result in particular crucially demonstrates that only a nonlinearity that can map to zero a large range of input values can effectively act as a filter, and for this reason the tanh function does not work well as a filter because its asymptotic values are $-1$ and $1$. However, it is important to note that the above analysis of FiLMv2 is largely empirically driven, and further work is needed to assess whether the improvements of FiLMv2 compared to FiLM hold more generally.

\subsection{Exponentially Weighted Mean Squared Error Loss (W-MSE)}
\label{methods:wmse}
As described in Section \ref{methods:spfd}, the goal of the surrogate model in the DSD framework is to select top molecule candidates which are then run through a classical docking algorithm. Thus, correct classification of top scoring data are far more important than correct classification of comparatively worse scoring data. Additionally, perhaps due to the presence of inductive biases in their algorithm design, current docking methods tend to produce much noisier scores for comparatively lower scoring molecules than higher scoring ones \cite{lyu_nature}. 
In order to address these two points, we introduce a weighted mean squared error (W-MSE) loss for use during training. This incentivizes our model to properly score the training samples with good docking scores. Additionally, the model is penalized less for bad predictions of bad scoring chemicals, as these labels may be comparatively noisier. Here, the loss of a single example is weighted by a negative exponential function  
of the label (docking score). Thus, labels associated with better docking scores (smaller scores) give a higher contribution to the overall loss. Explicitly, for a scalar valued model predicted score $z_i$ for molecule $m_i$, associated label $y_i$, mean squared error (MSE) loss function $l: \mathbb{R} \rightarrow \mathbb{R}$, and \textit{exponential coefficient} $\alpha$, we have: 
%
$loss = \sum_{i} e^{-\alpha y_i} \cdot l(z_i, y_i)$
%
%
Given that the most desirable (i.e. "best scoring") ligands correspond to those with the most negative scores, $\alpha$ hyperparameter in the range $[0, \infty)$, with $\alpha = 0$ corresponding to the case where no weighing occurs. In practice, we find that $\alpha = 0.8$ works well and thus use it for all models unless otherewise noted, though the effect of this hyperparameter is explored in more detail in the Appendix.

\subsection{Virtual Node}
\label{methods:virtual_node}
First introduced by \cite{virtual_node}, a so called \textit{virtual node} is a single node added to a graph with a connection to every other node in the given graph. The edge connections of the virtual node are given a special edge type to differentiate them from other nodes in the graph. It has been shown that the use of a virtual node improves performance by facilitating long distance connections between faraway nodes during the propogation phase of a GNN architecture\cite{virtual_node}. 
We introduce this feature to this application, adding a virtual node to each graph as a preprocessing step. Empirically we find that performance improves across all architectures tested. Further, the use of virtual nodes reduces reliance on computationally complex global pooling operations, thus reducing both train time and costly hyperaparamter tuning. The effects of the virtual node on this experiement are explored in ablation (Section \ref{ablation}).

\section{Experiments}
\label{experiments}

\begin{figure}
    \centering
    \subfigure[]{\includegraphics[width=0.38\textwidth]{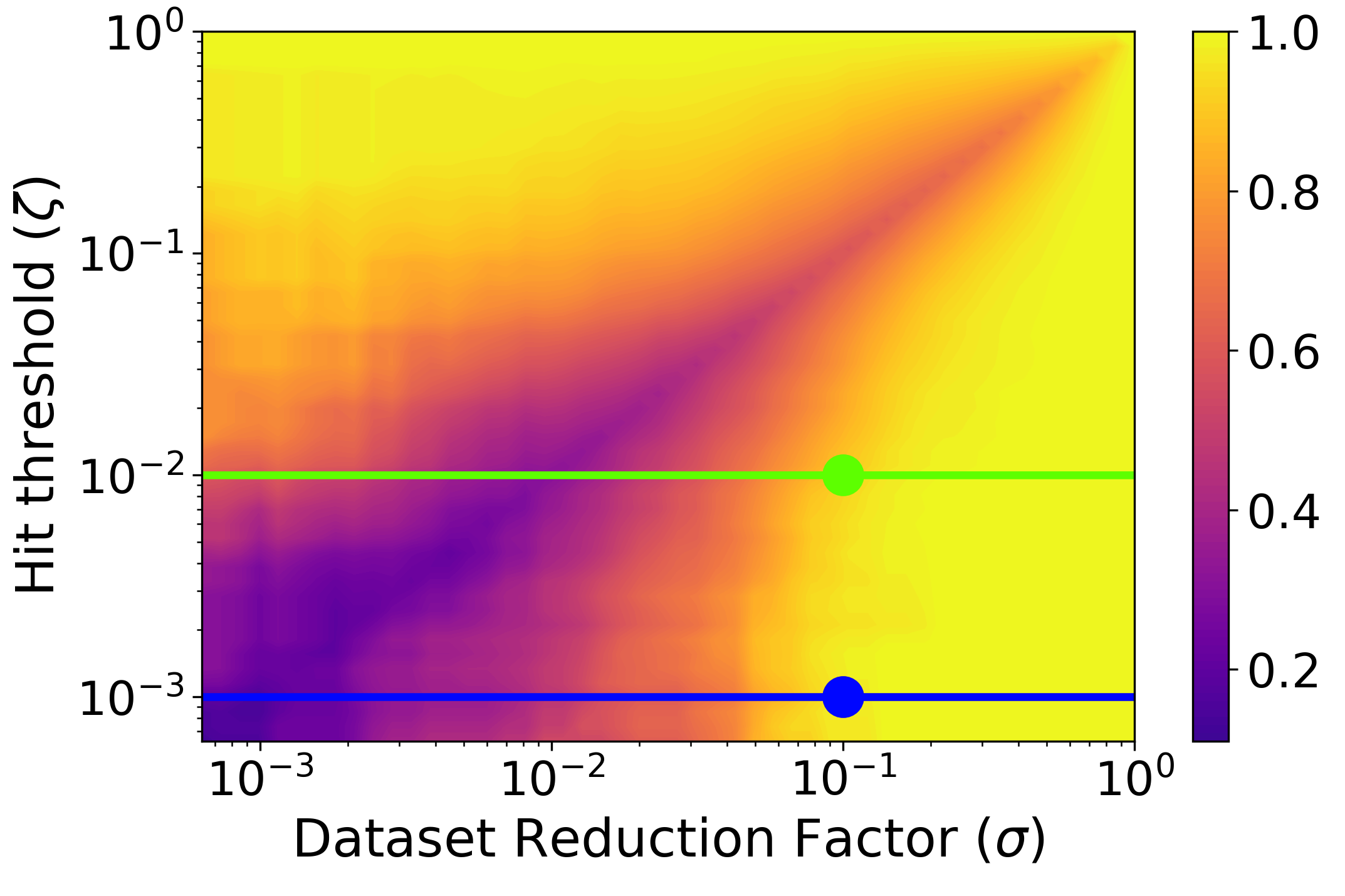}} 
    \subfigure[]{\includegraphics[width=0.35\textwidth]{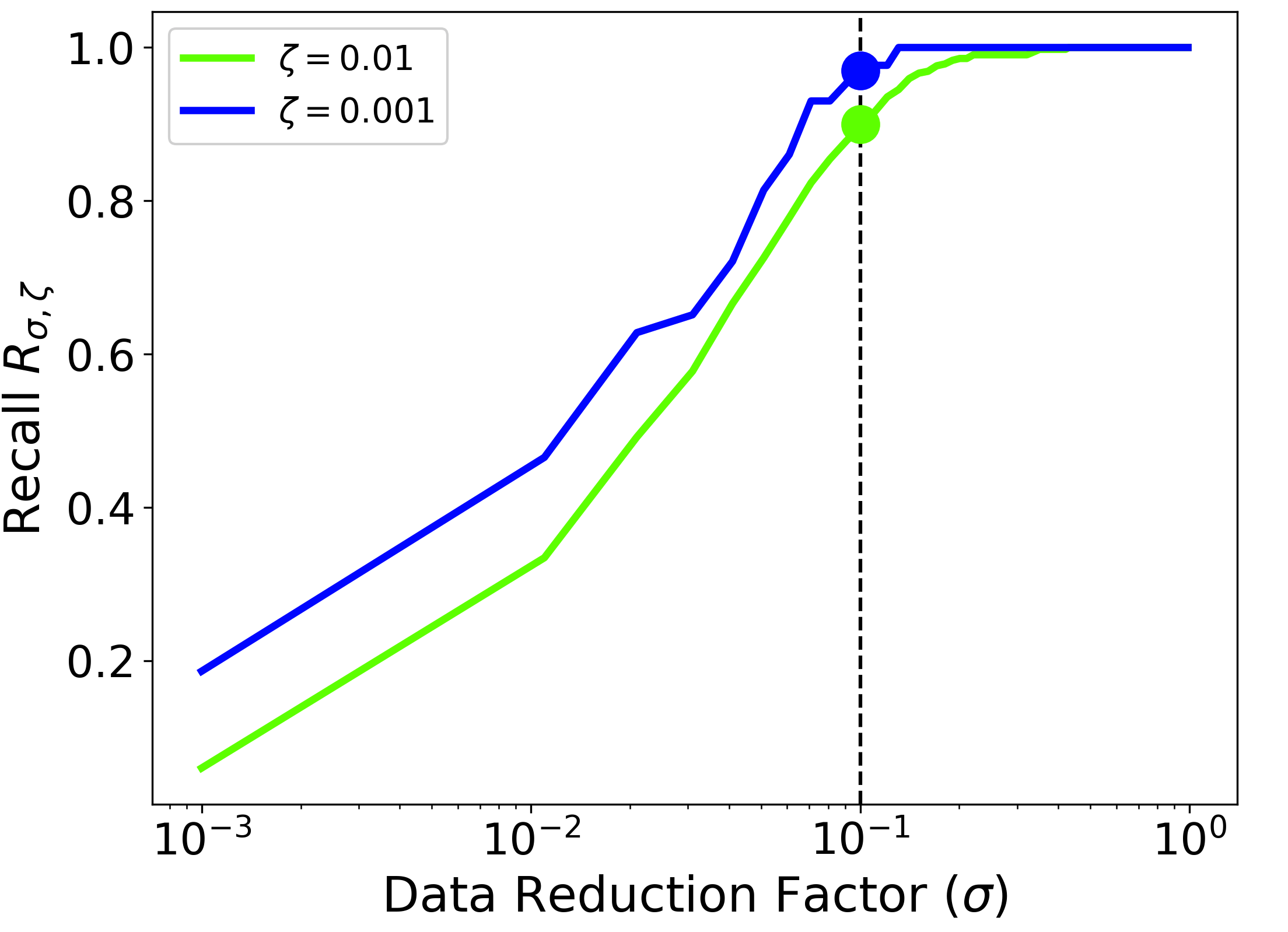}}
    \caption{(a) shows a RES plot of the FiLMv2 model, and superimposes the locations of AURTC$_{0.01}$ (green line), AURTC$_{0.001}$ (blue line), $R_{0.1, 0.01}$ (green dot), and $R_{0.1, 0.001}$ (blue dot). (b) shows the AURTC$_{0.01}$ and AURTC$_{0.001}$, and the locations of $R_{0.1, 0.01}$, $R_{0.1, 0.001}$ with respect to these curves.}
    \label{fig:metrics}
\end{figure}

We now describe and justify our experimental setup, 
defining the dataset, models, and metrics. 
We apply DSD to docking a virtual library of 128 million molecules with a single receptor target and we compare results of the FiLMv2 architecture with other state-of-the-art GNN models on a small held out test dataset, demonstrating the scalability and performance of the DSD framework. 
Finally, we ablate the important design elements of the work.


\subsection{Experimental Details}
\label{experiments:exp_details}

\paragraph{Datasets}
For experimentation, we use the ZINC dataset \cite{zinc} of 128 million virtual molecules with docking scores against the single D$_4$ dopamine receptor target, generated using the method described by \cite{lyu_nature}. Though to the best of our knowledge, no standard benchmark datasets yet exist for chemical docking, the ZINC dataset is one of the largest in size, diverse in molecule \textit{scaffold} representation, and contains other optimizations not found in earlier virtual libraries \cite{lyu_nature, zinc}. 
Additional details about the dataset description and preprocessing can be found in the Appendix.
%
For the majority of our experiments, we randomly sample 500,000 molecules with ground truth labels from the full dataset, 
corresponding to $\rho = 0.00313$ (see Section \ref{methods:spfd}), 
which are then split into 80-10-10 train-validation-test partitions. Models are tuned using the validation set and final results (in Section \ref{experiments:main_results}) are reported on the held out test partition. 
In Section \ref{experiments:main_results} we explicitly vary the $\rho$ parameter in order to demonstrate its effect on the overall framework. 
In Section \ref{experiments:inference}, labels for the full ZINC dataset are generated and used to evaluate model inference over the entire virtual library. 
%



\paragraph{Models}
We experiment implementing the surrogate model $f_{S}$ with 
four GNN convolution operations 
to use as a backbone for our architecture: the graph isomorphism operator (GIN) \cite{gin}, graph attention convolution operation (GATv2)~\cite{gatv2}, the feature-wise linear modulation operation (FiLM)~\cite{gnn_film}, and our novel FiLMv2 operation, which is discussed in detail in Section \ref{methods:filmv2}. 
Our full architecture consists of a backbone containing a hyperparameterized number of these graph operations, followed by a global mean pool layer and a MLP mapping the pooled features to a single regression score. Additionally, a single dropout layer and layer normalization layer are added between each graph convolution operation. Each of the four models were trained using the Adam optimizer in order to minimize the weighted mean squared error (W-MSE) loss presented in Section~\ref{methods:wmse} and tuned using standard hyperparameter tuning practices. Full details regarding this process can be found in the Appendix.

\paragraph{Metrics}

Given that the overall task of fitting a model with respect to the DSD framework is novel to the best of our knowledge, careful consideration was given to metrics used to evaluate $f_{S}$. 
For the purposes of evaluation, we can apply a hard threshold to both the ground truth scores and model predictions in order to convert the regression task into a binary classification problem. That is, following the logic in Section \ref{methods:spfd}, we can label the top $\zeta$ 
fraction of molecules scored by the classical docking algorithm $f_D$ 
as positive and otherwise negative. 
Similarly, we can threshold the top $\sigma$ 
fraction of molecules scored by the surrogate model $f_{S}$ 
as positive and otherwise negative. 
Thus, we can treat these labels as $y_{true}$ and $y_{pred}$ for the purposes of evaluating $f_{S}$ and use traditional binary classification metrics, like precision and recall.  
However, being functions of $\zeta$ and $\sigma$, the traditional metrics vary when different values of the two thresholds are selected and thus cannot offer a fair and comprehensive performance evaluation of the models. 
Thus, we introduce the {\it RES Score} metric to measure the overall performance of the models for all combinations of threshold values, the AURTC$_\zeta$ metric to measure the performance for a fixed $\zeta$, and provide the parametrized Recall values $R_{\sigma, \zeta}$ for specific values of $\zeta$ and $\sigma$. 
The next three paragraphs describe the RES Score, AURTC$_\zeta$, and R$_{\sigma, \zeta}$, respectively. Figure \ref{fig:metrics} visualizes the relationship between the three metrics. We also provide other typical classification metrics (e.g. AUROC, F-1, etc) for specific values of $\zeta$ and $\sigma$ in the Appendix. 


\paragraph{Regression Enrichment Surfaces and the \textit{RES Score}}
Regression Enrichment Surfaces (RES) have been introduced as a tool to mitigate some of the issues discussed above \cite{res}. RES can be conceptualized as a 3D surface parameterized by $\sigma$ and $\zeta$. For each $\sigma$, $\zeta$ $\in (0, 1)$, the value of RES is the recall of $y_{true}$ and $y_{pred}$, 
that is the fraction of all top-scoring molecules that were identified by the model. 
RES can be visualized using a heatmap, such as in Figure \ref{fig:filmv2_perf}, where axes are log scaled in order to highlight performance where $\sigma$ and $\zeta$ are close to zero. Additionally, a RES score can be calculated by taking the volume under the RES surface with log-scaled axes. 
This scalar summarizes the quality of the RES in a single number. 

\paragraph{Area Under Recall Threshold Curve (AURTC$_\zeta$)}
We define the \textit{Recall Threshold Curve} (RTC)
as a single dimensional slice of RES obtained by fixing a $\zeta$ value and expressing the recall as a function of $\sigma$. 
We define the area under this curve as AURTC$_\zeta$, which allows assessment of performance at particular values of $\zeta$. We report AURTC$_\zeta$ for all of our experiments at $\zeta = 0.01$ and $\zeta = 0.001$.

\paragraph{Parameterized Recall (R$_{\sigma, \zeta}$)}
Finally, we can follow the logic above and once again reduce the dimensionality of the 
AURTC$_\zeta$ to a single point by fixing $\sigma$ and obtaining the corresponding recall. 
We thus define parameterized recall, R$_{\sigma, \zeta}$, which reports a single recall value for a given $\sigma$ and $\zeta$. For all experiments, we report R$_{0.1, 0.01}$ and R$_{0.1, 0.001}$, which we believe represent reasonable typical parameters for most docking tasks. 

\subsection{Main Results}
\label{experiments:main_results}
\begin{figure}[h!]
    \centering
    \subfigure[]{\includegraphics[width=0.327\textwidth]{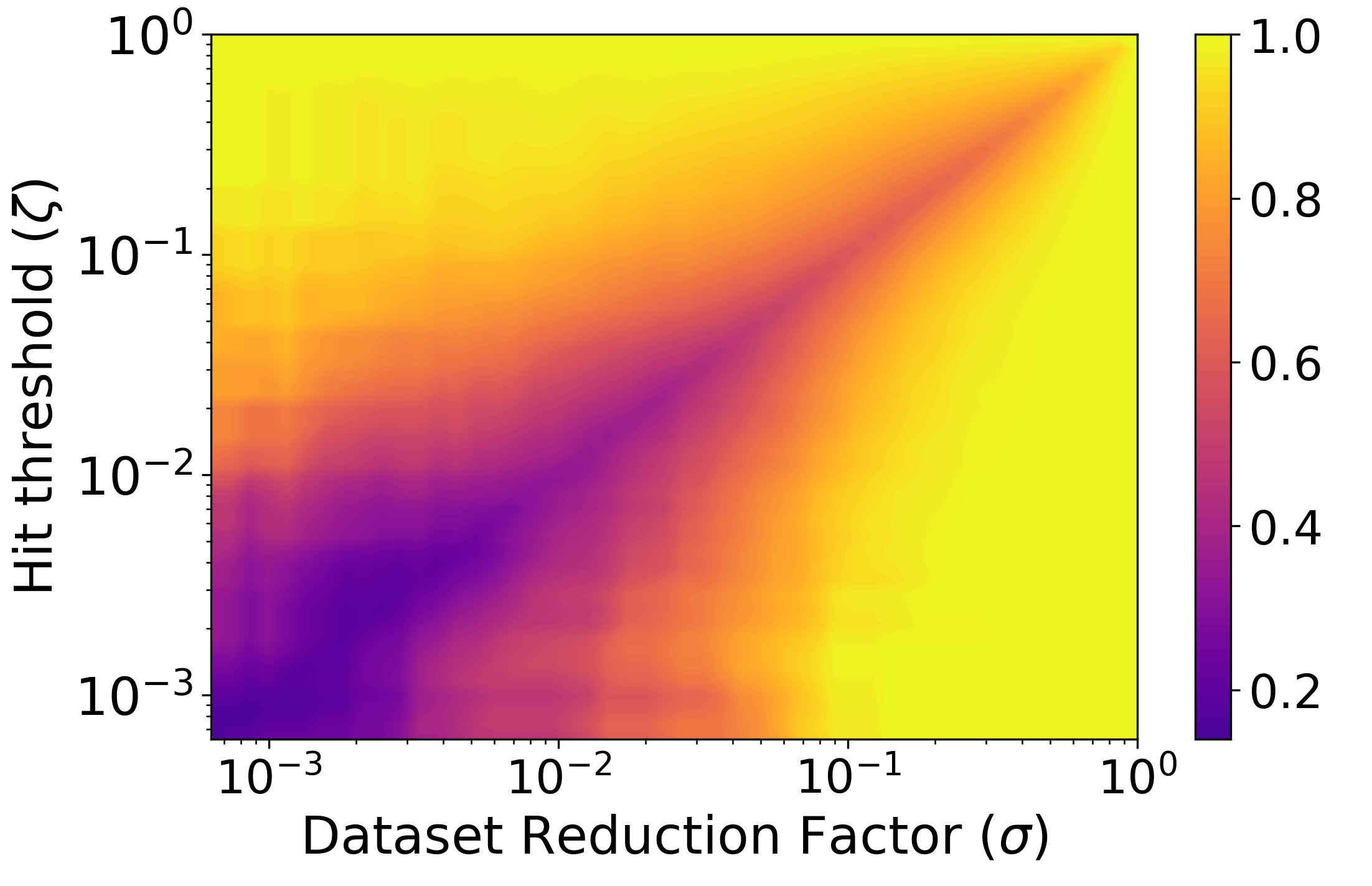}} 
    \subfigure[]{\includegraphics[width=0.329\textwidth]{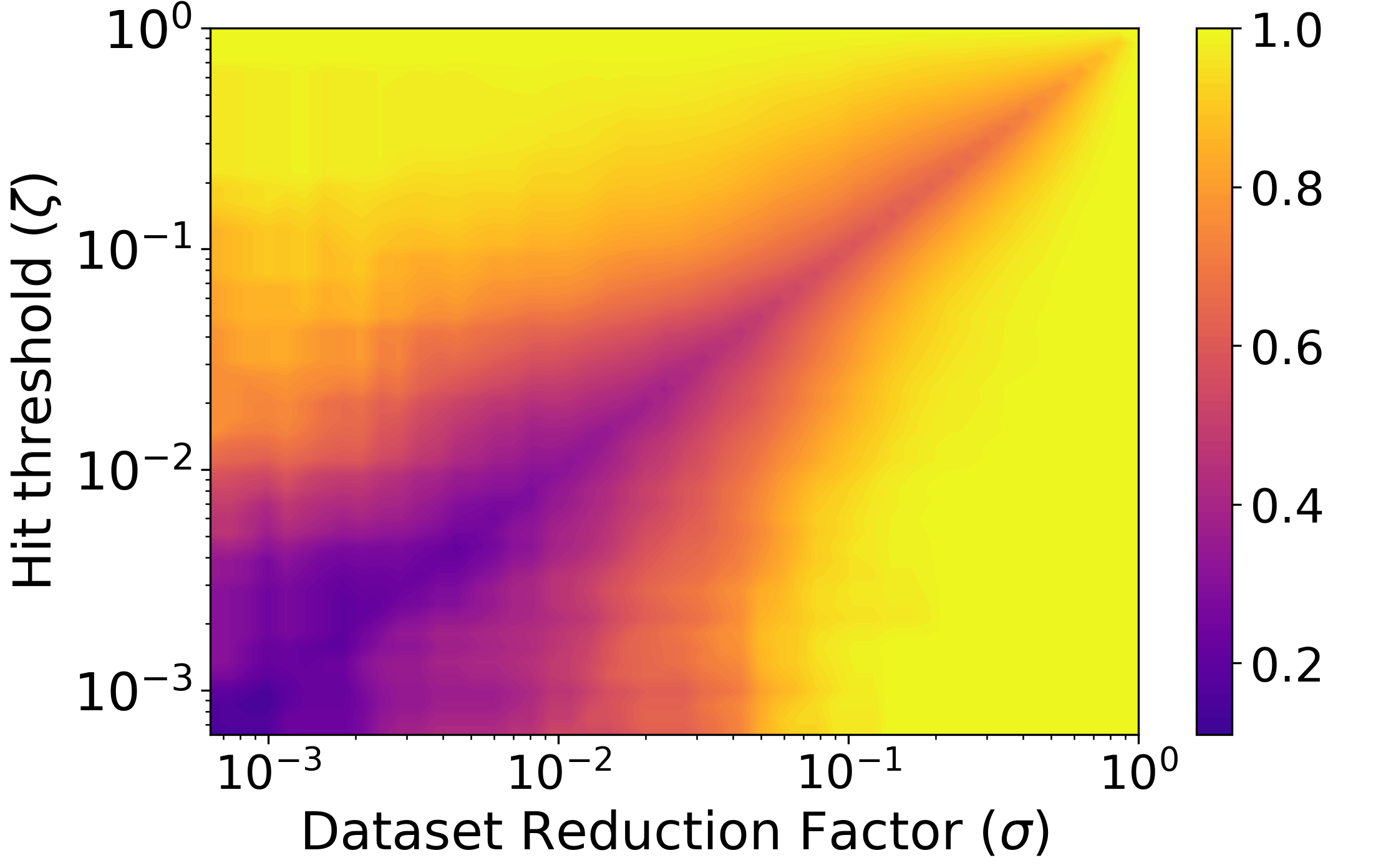}} 
    \subfigure[]{\includegraphics[width=0.327\textwidth]{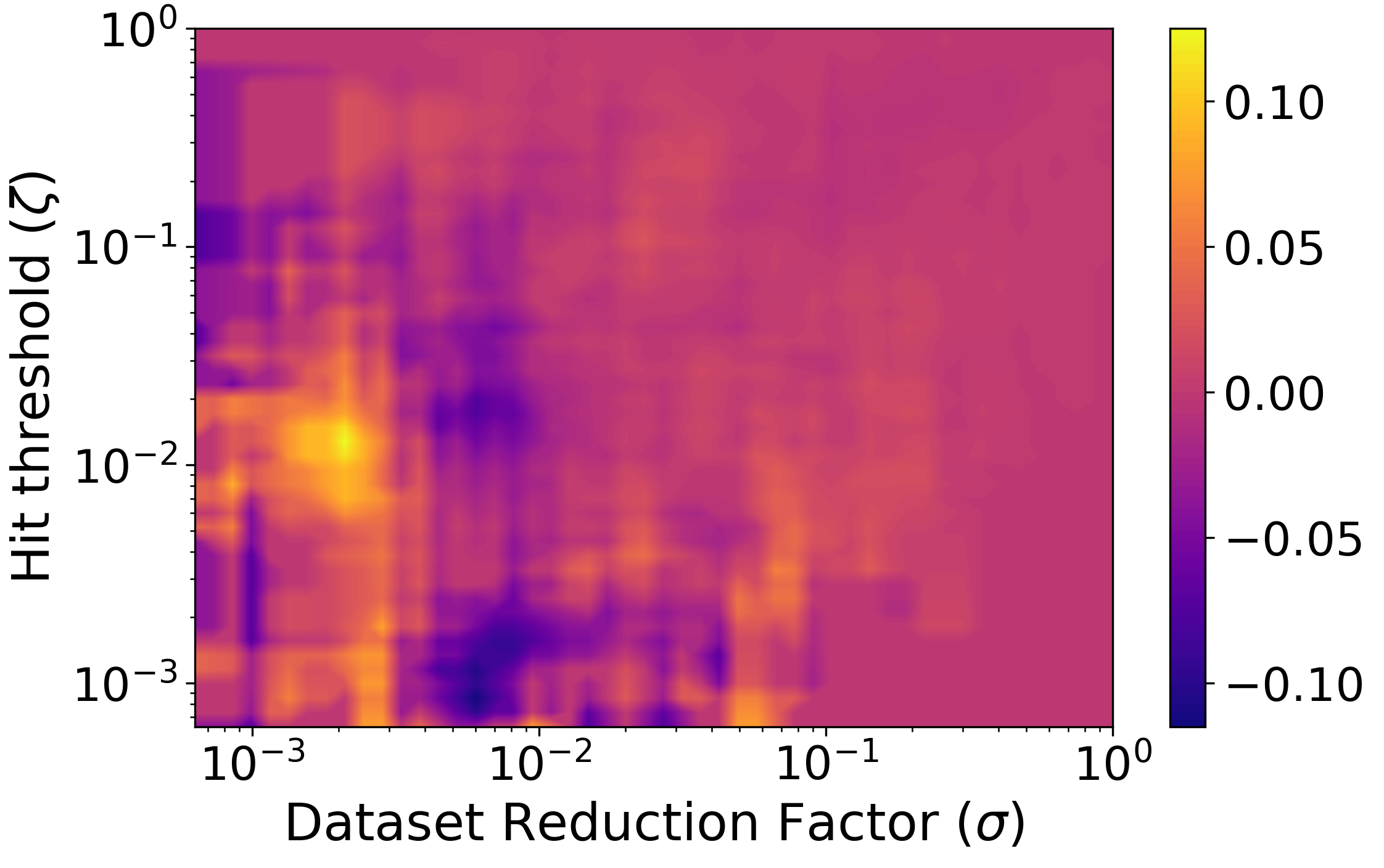}}
    \caption{
    RES plots of FiLM (a) and FiLMv2 (b) on a heldout test set. (c) shows the difference in recall values between FiLM and FiLMv2.}
    \label{fig:filmv2_perf}
\end{figure}

All four models were trained and hyperparameters tuned according to the details in Section~\ref{experiments:exp_details}. The performance on a held out test partition for each model is shown in Table~\ref{table:main_results}. We note that our novel FiLMv2 architecture outperforms all other models in all but one of the selected metrics, where perfomance is equivalent to the original FiLM architecture. Additionally, we find that 
AURTC$_\zeta$ \textit{improves} as $\zeta$ decreases (see Table~\ref{table:main_results} and Figure~\ref{fig:metrics}) demonstrating the model's success at accurately recalling molecules with the best scoring labels. 
Figure \ref{fig:filmv2_perf} shows the associated RES plots for FiLM and FiLMv2, as well as a heatmap showing the their difference. We note from this Figure that FiLMv2 additionally shows performance improvements in regions where both $\sigma$ and $\zeta$ are very small. As discussed previously, this is especially useful to the overall task and is indicative of FiLMv2's ability to adapt to the W-MSE loss function. Details about the final hyperparameters of each model, total number of model parameters, etc. can be found in the Appendix.

\begin{table}[ht!]
  \caption{Performance of FiLMv2 and other models from the literature as surrogate models on the ZINC dataset with labels generated from docking against the D$_4$ dopamine receptor. All results except FiLMv2 (128M inf.) are from a single held out test partition.}
  \label{table:main_results}
  \centering
  \begin{tabular}{lllllll}
    \toprule
    \cmidrule(r){1-2}
    Model & W-MSE & RES Score & AURTC$_{0.01}$ & AURTC$_{0.001}$ & R$_{0.1, 0.01}$ & R$_{0.1, 0.001}$\\
    \midrule
    GIN           & 0.402           & 0.742  & 0.938          & 0.951          & 0.798          & 0.804   \\
    GATv2         & 0.396           & 0.763  & 0.944          & 0.959          & 0.866          & 0.884    \\
    FiLM          & 0.389           & 0.768  & 0.946          & \textbf{0.965} & 0.890          & 0.968    \\
    FiLMv2 (ours) & \textbf{0.383}  & \textbf{0.773}  & \textbf{0.950} & \textbf{0.965} & \textbf{0.898} & \textbf{0.977}    \\
    \midrule
    FiLMv2 (128M inf.) & \textit{0.381} & \textit{0.775} & \textit{0.947} & \textit{0.967} & \textit{0.899} & \textit{0.972}  \\
    \midrule
    FiLMv2 (no W-MSE)& N/A & 0.758 & 0.932 & 0.960 & 0.838 & 0.954 \\
    FiLMv2 (no virt. node)& 0.402 & 0.760 & 0.944 & 0.963 &  0.878 & 0.976 \\
    \bottomrule
  \end{tabular}
\end{table}

While the majority of our experiments focus on a 500,000 molecule train-val-test dataset (i.e. $\rho=0.00313$), we vary the size of the dataset $D$ in order to observe its effect of performance. Thus, hyperparameter tuning on FiLMv2 architecture was conducted for several datset sizes. 
Table~\ref{table:more_main_results} in the Appendix shows the intuitive result that for each size tested, RES score and model size increase monotonically with the size of $D$. 


\subsection{Inference on 128 million virtual molecules}
\label{experiments:inference}
\label{experiments:inference_speed}
Here, we demonstrate the runtime and performance benefits of the DSD framework at scale by performing inference of the same 128 million ZINC dataset docked by \cite{lyu_nature}. To evaluate performance, the entire dataset (including the training samples) were given predictions by the final FiLMv2 model. The results of this inference is shown in Table \ref{table:main_results} as \textit{FiLMv2 (128M inf.)} and demonstrates that the results of evaluation on a small held out test set generalize well to the entire virtual library. 
We evaluate inference speed (i.e. \textit{screening throughput}) compared to traditional docking. We define $t_D$ to be the time required to dock this dataset using the traditional method in \cite{lyu_nature} and $t_{DSD} = t_{inf} + \sigma t_{D}$ to be the total time required for inference using our DSD framework. Here $t_{inf}$ is the time required to perform model inference on the full dataset and $\sigma t_{D}$ time required to dock the $\sigma N$ top candidates. For the given virtual library docked against the D$_{4}$ receptor, $t_{inf}$ was reported as 1,728 hours on 1,500 CPU cores. $t_{inf}$ for FiLMv2 architecture on a single A100 GPU is 9.167 hours. Thus, for $\sigma = 0.1$ the DSD method provides a $\frac{t_{D}}{t_{inf} + \sigma t_{D}} = \frac{1,728}{9.167 + 0.1\cdot 1,728}$ = 9.496x speedup, while recalling 97.7\% of the top 0.1\% of molecules. 
We note that this speed performance discussion is limited to inference time (known as \textit{docking throughput} in the literature \cite{docking_survey}), and does not include the one-time cost of generating a typically small labeled subset of the virtual library and training the model.

\subsection{Ablation Studies}
\label{ablation}
We ablate two major components of our model, the W-MSE loss presented in Section \ref{methods:wmse} and the virtual node presented in \ref{methods:virtual_node}. We thus train a FiLMv2 model with the same hyperparameters as the analogous FiLMv2 model in Section \ref{experiments:main_results}, but trained with vanilla MSE loss. We also train another FiLMv2 model without use of a virtual node in both training and inference. Table \ref{table:main_results} shows that both features result in improvement of the model across all reported metrics. W-MSE is especially effective in metrics that deal with the ability of the model to recall top scoring molecules, such as R$_{0.1, 0.01}$ and R$_{0.1, 0.001}$, improving performance by 0.05 and 0.023, respectively. This is consistent with our earlier discussions of W-MSE. 
The virtual node improves performance across metrics, for example increasing the RES Score and AUROC$_{0.01}$ by 0.013 and 0.006, respectively. However, it is important to note that the use of the virtual node can substantially increase training time, as a virtual node with edge connections to every other node increases the density of each graph. Thus, we additionally experiment with increasing model size when training on data without virtual node augmentation and find that, at a model size such that overall training time is similar to our final FiLMv2 model (with virtual node), the effect of the virtual node is less pronounced when compared to a FiLMv2 model of same size but trained with the virtual node augmentation. However, the overall performance of such models are worse than our final model. This substantiates our claim that use of virtual node both improves overall performance and allows use of smaller model sizes.

\section{Related Work}
\label{related_work}

There are generally two high level approaches to automated drug discovery, \textit{de-novo} design of molecules with desirable properties \cite{denovo_overview, barking, gan_denovo, generative_denovo, rl_generative_drugs} and screening based approaches \cite{docking_survey}. While \textit{de-novo} drug design, wherein generative models directly produce novel molecules with useful properties, shows significant promise, current approaches suffer from issues such as lack of molecule synthesizability, algorithmic instability, reduced molecule sizes, etc. \cite{generative_survey, barking}. Indeed, even \textit{de-novo} generated drugs may require or benefit from virtual screening, albeit at a comparatively smaller scale. 
Conversely, screening based approaches attempt to use a \textit{docking algorithm} in order to score molecules from a generally large virtual library of compounds for desirability by evaluating its binding affinity with a particular \textit{target} (also called \textit{receptor}, or \textit{ligand}) \cite{docking_survey}. Recently, substantial progress has been made to generate larger and more robust virtual libraries. Notably, \cite{lyu_nature} demonstrate successfully docking on a virtual library of 170 million molecules, and ultimately discovering a new class of inhibitors for the AMPC receptor. Crucially, they also show that hit rates of molecules decrease monotonically with docking score. This serves as one of the justifications for biasing our loss function to emphasize prediction of the best scoring molecules (See Section \ref{methods:wmse}). 
%
Recently, substantial work has focused on development of machine learning based surrogate models to accelerate docking. Some approaches focus on traditional feature based methods \cite{spfd, explainable}, while others apply language models directly on SMILES strings \cite{transformer, transformer2} or GNNs on graph representations of molecules \cite{lim2019predicting, emulating_docking_ml}. Of these GNN based approaches, most 
use trained models to infer docking scores directly (e.g. \cite{lim2019predicting}). 
Some existing screening methods \cite{dsd_1, dsd_2, spfd} do use a surrogate prefiltering approach, though they lack several of the key training and evaluation strategies proposed in our DSD workflow. 
Graph neural networks (GNNs) are generalizations of neural network architectures such as convolution that operate on non-euclidean data (i.e. graphs) \cite{gnn_survey}. At a high level, GNNs operate through \textit{message passing}, wherein the state of each node is iteratively updated using a function of the state of its neighbor nodes. Recently, substantial progress has been made in designing GNN general architectures that are analogous to common DL architectures such as convolution \cite{gnn_film, gcn, gcn2, sage, pan, pdn, signed_conv}, VAEs \cite{gvae, arvae}, attention \cite{gat, gatv2, hyper, abgnn, rgan, neighbor_attn, het_gan, het_gan2}, and others \cite{infomax, node2vec, autoRNN, gunet}. Additionally, substantial work has been made in creating GNN architectures suitable for specific applications such as for modeling quantum interactions \cite{schnet} and molecular property prediction \cite{dime, attentivefp}, though we generally avoid these models in our experimentation as they may inadvertently introduce implicit biases to our framework. 


\section{Conclusions}
\label{conclusions}
In this work, we present \textit{Deep Surrogate Docking} (DSD), a workflow which accelerates traditional chemical docking through the use of surrogate models. 
We also demonstrate the viability of DSD in the drug discovery domain by using it to dock 128 million molecules in the ZINC dataset against the D$_4$ dopamine receptor. We show that DSD substantially accelerates docking throughput with minimial error. For example, in one such setup, we are able to speedup a standard docking algorithm by 9.496x, while recalling $>97\%$ of top 0.1\% molecules. Moreover, we show that the DSD workflow allows scientists flexibility to tradeoff inference time and hit accuracy by varying a single data reduction parameter $\sigma$.
We also present a novel GNN architecture, FiLMv2, which we show outperforms other models from the literature in the example docking task above. FiLMv2 is implemented as a simple modification to the existing Feature-wise linear modulation (FiLM) architecture and  attains considerably more accurate and stable performance by allowing the model to more easily filter out irrelevant information from data. 
This work demonstrates the viability of the DSD workflow with a single large scale example. We leave the analysis of generalizability to other virtual molecule libraries and receptors to future work. 

\begin{ack}
This research used resources of the Argonne Leadership Computing Facility, which is a DOE Office of Science User Facility supported under Contract DE-AC02-06CH11357. We would like to thank Venkatram Vishwanath for useful discussion, feedback, and support. 
\end{ack}

\bibliography{refs}

\appendix
\section*{Appendix} 

\section{Additional Details Regarding Chemical Docking for Drug Discovery}
\label{appendix_a}
Here, we discuss and clarify details about classical docking algorithms which are relevant to this work. 

\paragraph{Overview of Classical Docking Algorithms}
At a high level, docking algorithms are functions that (1) sample and (2) score molecules against a ligand/receptor. This scoring function estimates the binding affinity of a candidate molecule against the receptor. As discussed in the main text, they are thus often used to screen molecules (i.e. potential drug candidates) against a specific ligand. The binding affinity calculation is acheived by simulating the interaction of a given molecule with the ligand at several orientations (that is, \textit{poses}) \cite{docking_algos}. However, most docking algorithms only succeed in ranking the relative binding affinities of a set of molecules and either do not aim to, or are not successful in, predicting the true binding affinity of the molecule-lingand pair\cite{docking_algos}. It is also important to note that recently \cite{lyu_nature} showed that the \textit{hit-rate}, or quality of docking prediction, decreases monotonically with the docking score. Thus, many modern screening workflows use docking as a prefilter, and then use domain specific heuristics to further narrow down the set of molecules for \textit{in-vitro} synthesis \cite{lyu_nature, docking_survey, spfd}. This further motivates the use of surrogate models as drop-in prefilters in the automated drug discovery pipeline.

\paragraph{Pose Estimations}
As discussed above, \textit{pose estimations} are calculated by docking algorithms, and are thus output by most docking software \cite{docking_algos}. However, for simplicity, we define the docking function $f_{D}$ in the main text as only outputting a docking score. In practice, both the score and pose can be recovered from $f_D$. However, the pose is not relevant to the DSD workflow.

\paragraph{Lack of Parallelizability}
Due to the sequencial nature of the docking process, there is currently no good way to parallelize the docking of a single molecule, and thus docking algorithms are generally CPU bound \cite{lyu_nature, docking_algos}. In our opinion, this further motivates the need for accurate surrogate models in order to accelerate the automated drug discovery process.

\section{Experimental Details}
\label{appendix_b}
Here, we detail our preprocessing and training setup, hyperparameter tuning strategy, provide details about the hardware system used, and provide additional metrics for the performance of the top models for each of the four tested architectures listed in Table 1 one of the main text. 

\paragraph{Preprocessing Details}
ZINC data is provided in \textit{Simplified Molecular Input Line Entry System}, or \textit{SMILES}, format, a common string representation of molecules \cite{smiles}. We use standard techniques~\cite{emulating_docking_ml} to convert each SMILES string into a (two-dimensional) graph representation for use by a GNN. Specifically, we create a graph representation of each molecule where the node represents a single atom in a molecule and contains nine features: atomic number, chirality, degree, formal charge, number of associated hydrogen bonds, number of radical electrons, hybridization, whether it is aromatic, and whether it is contained in a ring. Edges represent the bond between atoms in a molecule and are encoded with three features: the bond type, stereochemical information, and whether it is conjugated. More information about the technical details of these features, as well as their \textit{in-silico} representations can be found in the RDKit documentation \cite{landrum2013rdkit}, which was the tool used to perform this preprocessing. Additionally, as described in Section 2.4 of the main text, a virtual node with a distinct \textit{virtual} edge type to every other node in the graph was added to each molecule during preprocessing. Molecules with NaN docking values where dropped, as the docking algorithm used by \cite{lyu_nature} represents either noisy or poor performing molecules with NaNs. Finally, all docking labels were standardized by the mean and standard deviation of the training set label distribution. 

\paragraph{Model Training and Hyperparameter Tuning}
As described in Section 3 of the main text, we experiment with implementing four GNN convolution operations to use as a backbone for our architecture: the graph isomorphism operator (GIN)\cite{gin}, graph attention convolution operation (GATv2) \cite{gatv2}, the feature-wise linear modulation operation (FiLM) \cite{gnn_film}, and our novel FiLMv2 operation, which is discussed in detail in Section 2.2 of the main text. For each graph operation, we stack a hyperparameterized number of layers of said operation in order to create the backbone of the model. A layer normalization layer, dropout layer, and ReLU non-linearity are applied between each of these layers. After this backbone, a global mean pool aggregates the node representations and a single MLP layer maps this aggregated representation to a single regression score. In order to simplify the hyperparameter tuning process, a single hidden dimension size and dropout amount is chosen for each of the layers. Thus, the hyperparameters are the size of this hidden layer, the number of layers in the backbone, and the amount of dropout. For all models, hyperparameter tuning was run, via grid search, on hidden layer sizes of $\{2, 4, 8, 16, 32, 64, 128, 256\}$, number of layers of $\{1, 2, 4, 5, 6\}$, and dropout values of $\{0.0, 0.1, 0.15, 0.2, 0.3\}$. For hidden layer sizes, additional tuning was done in multiples of four at a local neighborhood around the initial hidden layer size chosen. For example, if the best performing hidden layer size in the initial grid search was 64, additional testing was done for $\{52, 56, 60, 64, 68, 72, 76, 80\}$. Each model was trained for 1,000 epochs, and the epoch where validation loss was best was selected as the final model. All models were trained with the Adam optimizer with a learning rate of 0.001 and a batch size of 512. The final hyperparameter values for the best model for each of the four architectures (corresponding to the first four rows of Table 1 in the main text) are shown in Table \ref{table:more_main_results}. 

\paragraph{Hardware Details}
Training was conducted using an internal cluster of 8 NVIDIA DGX A100 GPUs (40GB memory), using Pytorch's built-in distributed data parallel (DDP) tool. Inference was conducted using a single NVIDIA DGX A100 GPU (40GB memory). Total computation required to train the best performing model is 29.421 GPU hours. Total computation used to perform inference on the total dataset of 128 million molecules is 9.167 GPU hours. 

\paragraph{Additional Metrics for Final Models}
As discussed in Section 3.1 of the main text, it is possible to convert the DSD task to a classification problem by fixing the parameters $\sigma$ and $\zeta$. Table \ref{table:more_main_results} also reports the AUROC and F-1 scores for each of the models for $\sigma=0.1$ and $\zeta=0.001$. Note that as AUROC by nature varies the classification threshold (i.e. $\sigma$), a fixed $\sigma$ is not used to calculate the AUROC.

\begin{table}[h]
  \caption{Hyperparameter values and additional metrics for the best performing models of each architecture. These models correspond to the first four rows of Table 1 in the main text. RES Score is repeated in this table to facilitate easier reference.}
  \label{table:more_main_results}
  \centering
  \begin{tabular}{llllllll}
    \toprule
    \cmidrule(r){1-2}
    Model & Hidden Layer Size & \#Layers & Dropout & \#Params. & RES Score & AUROC & F-1\\
    \midrule
    GIN           & 120            & 4       & 0.1   & 132,961          & 0.742          & 0.979       & 0.998    \\
    GATv2         & 64             & 4       & 0.1   & 105,537          & 0.763          & 0.986       & \textbf{0.999}    \\
    FiLM          & 64             & 4       & 0.1   & \textbf{102,977} & 0.768          & 0.985       & \textbf{0.999}    \\
    FiLMv2 (ours) & 64             & 4       & 0.1   & \textbf{102,977} & \textbf{0.773} & \textbf{0.987}       & \textbf{0.999}    \\
    \bottomrule
  \end{tabular}
\end{table}

\paragraph{Error Bars} As part of our experimentation, we retrained the best performing models for all four architectures 5 times, varying the random seeds used each time. For all models, the range of W-MSE was $<0.0002$ and the range of the RES Score was $<0.0004$. Thus, we conclude that the results obtained in this work are not significantly affected by stochasticity in the training process.

\section{Additional Details on the Effects of Dataset Size}
\label{appendix_c}
Section 3.3 describes the phenomenon that as the dataset size used for training is increased, so does the RES Score and number of parameters of the best tested model. In our experiments, we performed grid search on the FiLMv2 model, the best performing architecture from Section 3.2. Specifically, we used the same assumptions for model architecture as in \ref{appendix_b} and tested hidden dimension sizes of $\{4, 8, 16, 32, 64, 128\}$, number of convolutional layers of $\{1, 2, 4, 5, 6\}$, and dropout of $\{0.1, 0.15, 0.2\}$. After finding the optimal model associated with the combination the hyperparameters above, we performed a smaller search around the chosen hidden dimension in order to further tune this hyperparameter. Each model was trained for 1,000 epochs, and the epoch where validation loss was best was selected as the final model. Table \ref{table:ds_size} shows the RES score and number of parameters for the final model chosen for testing after hyperparameter tuning. Please note that the dataset size refers to the total dataset size and, as described in 3.1, this number is further broken down into a 80-10-10, train-val-test split. For example, the first row of Table \ref{table:ds_size} has a dataset of 5,000, of which $5,000 \cdot 0.8 = 4,000$ examples were used for training.  Table \ref{table:ds_size_hp} shows the chosen hyperparameters for each of these models. 

\begin{table}[h]
  \caption{RES and model sizes for different sizes of D.}
  \label{table:ds_size}
  \centering
  \begin{tabular}{lll}
    \toprule
    \cmidrule(r){1-2}
    Dataset Size & RES Score & \#Params. \\
    \midrule
    5,000     & 0.594 & 7,313 \\
    50,000    & 0.724 & 26,913 \\
    100,000   & 0.756 & 58,801 \\
    500,000   & 0.773 & 102,977 \\
    2,000,000 & 0.781 & 600,193 \\
    \bottomrule
  \end{tabular}
\end{table}

\begin{table}[h]
  \caption{Hyperparameters of each of the models in Table \ref{table:ds_size}}
  \label{table:ds_size_hp}
  \centering
  \begin{tabular}{llll}
    \toprule
    \cmidrule(r){1-2}
    Dataset Size & Hidden Layer Size & \#Layers & Dropout \\
    \midrule
    5,000     & 16         & 4       & 0.1        \\
    50,000    & 32         & 4       & 0.1        \\
    100,000   & 48         & 4       & 0.1        \\
    500,000   & 64         & 4       & 0.1        \\
    2,000,000 & 128        & 6       & 0.1        \\
    \bottomrule
  \end{tabular}
\end{table}


\section{Additional FiLMv2 Experiments}
\label{appendix_d}
Section 2.2 of the main text describes the architecture of our novel FiLMv2 convolution in detail. There, three main empirical claims are used for discussion:
\begin{enumerate}
    \item The nonlinearity $sigma$ in Eq. 1 of the main text can be removed as it does not improve the model’s performance.
    \item Adding a nonlinearity to the source term ($(\boldsymbol{W}_\alpha \boldsymbol{h}_{u}^{t})$) does not further improve the model performance.
    \item Using a tanh instead of ReLU nonlinearity does not bring any improvement with respect to the original FiLMv1.
\end{enumerate}

The experiments conducted that led to these conclusions are detailed below for completeness. We define \textit{FiLMv2 (tanh)} as the FiLMv2 model defined in Section 2.2 of the main text, but with all instances of ReLU non-linearity replaced with tanh non-linearity. We also define \textit{FiLMv2 (source act.)} as the FiLMv2 model defined in Section 2.2 of the main text, but with an additional ReLU non-linearity added to source term ($(\boldsymbol{W}_\alpha \boldsymbol{h}_{u}^{t})$). \textit{FiLM} and \textit{FiLMv2} are as defined in the main text. 

Table \ref{table:film_exp} shows the results of testing each of these models on the 500k train-val-test set used for training the main models. Each model was created by using one of the four convolution operations described above (FiLM, FiLMv2, FiLMv2 (tanh), FiLMv2 (source act.)) and building a backbone of four layers with a hidden dimension size of 64. All other details, such as dropout and layer norm, global pooling, and MLP, etc. are the same as in \ref{appendix_b}. Thus, all models have the same number of parameters, and are trained with the same seed on the same dataset.

\begin{table}[h]
  \caption{Performance comparison of FiLMv2 with FiLM and FiLMv2 variants.}
  \label{table:film_exp}
  \centering
  \begin{tabular}{lllllll}
    \toprule
    \cmidrule(r){1-2}
    Model & W-MSE & RES Score & AURTC$_{0.01}$ & AURTC$_{0.001}$ & R$_{0.1, 0.01}$ & R$_{0.1, 0.001}$\\
    \midrule
    FiLM                 & 0.389                & 0.768           & 0.946       & 0.965             & 0.890          & 0.968    \\
    FiLMv2               & 0.383                & 0.773           & 0.950       & 0.965             & 0.898          & 0.977    \\
    FiLMv2 (tanh)        & 0.388                & 0.765           & 0.947       & 0.966             & 0.892          & 0.967    \\
    FiLMv2 (source act.) & 0.385                & 0.770           & 0.949       & 0.965             & 0.897          & 0.974    \\
    \bottomrule
  \end{tabular}
\end{table}

\section{W-MSE Loss and the \textit{alpha} Hyperparameter}
\label{appendix_e}
Section 2.3 in the main text describes the weighted mean square error (W-MSE) used in order to increase the recall of top scoring molecules. Here, we show the results for training the top performing FiLMv2 model from Section 3.2 of the paper at several values of $\alpha$, the hyperparameter responsible for the degree to which each molecule is weighted (See Section 2.3 for more details). Table \ref{table:exp_tests} shows the results of such experiments. Since varying $\alpha$ changes the value of the overall loss for a constant dataset and model, the absolute W-MSE loss when varying $\alpha$ may not necessarily be indicative of downstream performance. Thus, Table \ref{table:exp_tests} also shows performance of each model for all the metrics used for evaluation in Table 1 of the main text. We see that $\alpha = 0.8$ matches or beats performance on all but one of these metrics and thus use $\alpha=0.8$ for all other experiments.

\begin{table}[h]
  \caption{W-MSE and several metrics for the best peforming FiLMv2 model trained with different W-MSE weighing schemes (i.e. different $\alpha$ vales).}
  \label{table:exp_tests}
  \centering
  \begin{tabular}{lllllll}
    \toprule
    \cmidrule(r){1-2}
    $\alpha$ & W-MSE & RES Score & AURTC$_{0.01}$ & AURTC$_{0.001}$ & R$_{0.1, 0.01}$ & R$_{0.1, 0.001}$\\
    \midrule
    0           & 0.401           & 0.758           & 0.932          & 0.960          & 0.838          & 0.954            \\
    0.2         & 0.358           & 0.764           & 0.942          & 0.964          & 0.869          & 0.953             \\
    0.5         & \textbf{0.348}  & \textbf{0.773}  & 0.947          & 0.965          & 0.893          & \textbf{0.977}    \\
    0.8         & 0.383           & \textbf{0.773}  & \textbf{0.950} & 0.965          & \textbf{0.900} & \textbf{0.977}    \\
    1.0         & 0.433           & 0.770           & 0.949          & 0.966          & 0.893          & \textbf{0.977}    \\
    1.2         & 0.499           & 0.762           & \textbf{0.950} & \textbf{0.970} & 0.893          & 0.953              \\

    \bottomrule
  \end{tabular}
\end{table}

\section{Tools}
\label{appendix_f}

\paragraph{Software Tools}
Substantial portions of this project were implemented in the Pytorch \cite{pytorch} and Pytorch Geometric \cite{pyg} libraries, both of which are open source (BSD and MiT licenses, respectively) software tools. Each tool was utilized with adherence to their licensing conditions. We used Weights \& Biases \cite{wandb}, which is free for academic use (MiT license) for experiment tracking and hyperparamter sweeps. Additionally, several other Python libraries were used for various auxiliary tasks ranging from preprocessing to utility tasks. A full list of these tools can be found in our code's \texttt{requirements.txt} file. Again, each tool was used according to their respective library's licensing conditions. 

\paragraph{Data}
As discussed in the main text, data from the ZINC \cite{zinc} dataset was used. ZINC is a free and publicly available dataset, and was used according to their licensing policies. Additionally, labels for this data was provided by the docking conducted by \cite{lyu_nature}, and was used in this project with permission from the authors.



\end{document}